\documentclass[10pt,letterpaper]{article}

\usepackage{times}
\usepackage{epsfig}
\usepackage{graphicx}
\usepackage{amsmath}
\usepackage{amssymb}

\newcommand{\btheta}{\boldsymbol\theta}
\newcommand{\bx}{{\bf x}}
\newcommand{\argmax}{\operatornamewithlimits{argmax}}

\newcommand{\etal}{\mbox{\emph{et al.\ }}}
\newcommand{\ie}{\mbox{\emph{i. e.\ }}}

\usepackage[pagebackref=true,breaklinks=true,letterpaper=true,colorlinks,bookmarks=false]{hyperref}

\begin{document}

\title{Learning Analysis-by-Synthesis for 6D Pose Estimation in RGB-D Images}

\author{Alexander Krull\\
\and
Eric Brachmann\\
\and
Frank  Michel\\
\and
Michael Ying  Yang\\
\and
Stefan  Gumhold\\
\and
Carsten  Rother\\
\\
TU Dresden,
Dresden, Germany\\
}

\maketitle

\begin{abstract}
Analysis-by-synthesis has been a successful approach for many tasks in computer vision, such as 6D pose estimation of an object in an RGB-D image which is the topic of this work. The idea is to compare the observation with the output of a forward process, such as a rendered image of the object of interest in a particular pose. Due to occlusion or complicated sensor noise, it can be difficult to perform this comparison in a meaningful way. We propose an approach that ``learns to compare'', while taking these difficulties into account. This is done by describing the posterior density of a particular object pose with a convolutional neural network (CNN) that compares an observed and rendered image. The network is trained with the maximum likelihood paradigm. We observe empirically that the CNN does not specialize to the geometry or appearance of specific objects, and it can be used with objects of vastly different shapes and appearances, and in different backgrounds. Compared to state-of-the-art, we demonstrate a significant improvement on two different datasets which include a total of eleven objects, cluttered background, and heavy occlusion.
\end{abstract}

\section{Introduction}

\begin{figure}[!ht]
\begin{center}

\includegraphics [width=0.60\textwidth]{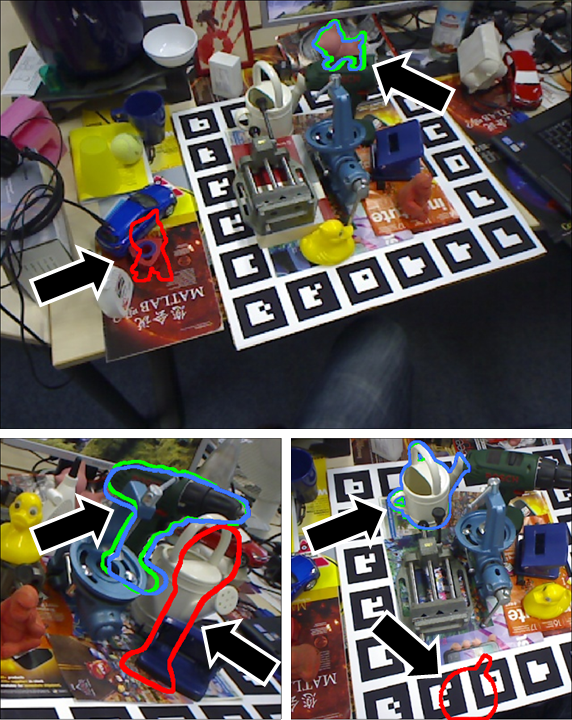}
\end{center}
   \caption{Three pose estimation results from the \emph{occlusion dataset} from \cite{brachmann2014} and \cite{hinterstoisser2012accv}. Arrows indicate the positions of estimated and ground truth poses. The green silhouette indicates the ground truth pose, the blue silhouette corresponds to our estimated pose. Red indicates the pose estimate from \cite{brachmann2014}. The marker board served only for ground truth annotation.}
\label{fig:pose-result}
\end{figure}

Tremendous effort has focused on the tasks of object instance detection and pose estimation in images and videos. 
In this paper we consider the pose estimation in a single RGB-D image, as shown in Fig.~\ref{fig:pose-result}.
Given the extra depth channel, it becomes feasible to extract the full 6D pose (3D rotation and 3D translation) of object instances present in the scene.
Pose estimation has important applications in many areas, such as robotics \cite{Martinez_Torres_2010,zhu2014grasping}, medical imaging \cite{ralovich_2014}, and augmented reality \cite{hagbi_2011}.
Recently, Brachmann \etal \cite{brachmann2014} achieved state-of-the-art results by adapting analysis-by-synthesis approach for pose estimation in RGB-D images. They use a random forest \cite{breiman_2001} to obtain pixelwise dense predictions.
Building upon the system of \cite{brachmann2014}, we propose a novel method to learn to compare in the analysis-by-synthesis framework. We use a convolutional neural network (CNN) inside a probabilistic context to achieve this.

Analysis-by-synthesis has been a successful approach for many tasks in computer vision, such as object recognition \cite{Hejrati_CVPR_2014}, scene parsing \cite{Isola_SceneCollaging_2013}, pose estimation and tracking \cite{GallRS_2008}.
A forward synthesis model generates images from possible geometric interpretations of the world, and then selects the interpretation that best agrees
with the measured visual evidence. 
In particular for pose estimation, the idea is to compare the observation with the output of a forward process, such as a rendered image of the object of interest in a particular pose.  
When attempting pose estimation in RGB-D images, comparing for analysis-by-synthesis is nontrivial due to occlusion or complicated sensor noise. 
There are for example areas with no depth measurements in Kinect or poor IR-reflectance.

\subsection{Contributions}

\begin{itemize}
	\item We achieve considerable improvements over state-of-the-art methods of pose estimation in RGB-D images with heavy occlusion.
  \item To the best of our knowledge, this work is the first to utilize a convolutional neural network (CNN) as a probabilistic model to learn to compare rendered and observed images.
  \item We observe that the CNN does not specialize to the geometry or appearance of specific objects, and it can be used with objects of vastly different shapes and appearances, and in different backgrounds.
\end{itemize}

The paper is organized as follows.
Section~\ref{sec:RelatedWork} provides an overview of related work.
Our proposed approach is described in Sec.~\ref{sec:methods}.
In Sec.~\ref{sec:exp} we present evaluation of our method compared to the state-of-the-art on two datasets.
We conclude the paper in Sec.~\ref{sec:con}.

\section{Related Work}
\label{sec:RelatedWork}

A large body of work in computer vision has focused on the problem of object detection and pose estimation, including instance and category recognition, rigid and articulated objects, and coarse (quantized) and accurate (6D) poses. Pose estimation has been an active topic, ranging from  template-based approaches \cite{hinterstoisser2012accv,dantone_pose_2014}, sparse feature-based approaches \cite{Martinez_Torres_2010}, and dense approaches \cite{Shotton_2013,brachmann2014}.
In the brief review below, we focus on techniques that specifically address CNNs and analysis-by-synthesis.

\noindent{\bf CNNs.}
are driving advances in computer vision in recent years, such as image classification \cite{krizhevsky_cnn_2012}, detection \cite{zhang_rcnn_2014}, recognition \cite{agrawal_nn_2014,oquab_cnn_2014}, semantic segmentation \cite{long_shelhamer_fcn_2015}, pose estimation \cite{toshev_pose_2014}.
CNNs have shown remarkable performance in the large-scale visual recognition challenge (ILSVRC2012). 
The success of CNNs is attributed to their ability to learn rich feature representations as opposed to hand-designed features used in previous image classification methods. 
In \cite{gupta_ECCV14}, rich image and depth feature representations have been learned with CNNs to detect objects in RGB-D images. 
In \cite{brox_2015}, CNNs are used to generate an RGB image given the set of 3D chair models, the chair type, viewpoint and color.
Very recent work from Gupta \etal \cite{Gupta_pose_2015} uses object instance segmentation output from \cite{gupta_ECCV14} to infer 3D object pose in RGB-D images.
Another CNN is used to predict the coarse pose of the object. This CNN is trained using pixel normals in images containing rendered synthetic objects.
This coarse pose is used to align a small number of prototypical models to the data, and place the model that fits the best into the scene.
Different from above approaches, we use a CNN as a probabilistic model to compare rendered and observed images. The output of our CNN is the energy value, while in \cite{Gupta_pose_2015} the output of the CNN is the object pose.
In \cite{Chopra_2005}, a similarity metric is learned. The learning process minimizes a discriminative loss function.
A CNN with \textit{siamese} architecture is used for mapping two face feature spaces. Similarly, in \cite{wohlhart2015learning} Wohlhart and Lepetit train a CNN to map image patches to a descriptor space, where pose estimation and object recognition is solved using the nearest neighbor method.  
Our framework is probabilistic. The posterior distribution of the pose is modelled as a Gibbs distribution with a CNN as energy function.
Zbontar and LeCun \cite{LeCun2014} train a CNN to predict how well two image patches match and use it to compute the stereo matching cost. 
The cost is minimized by cross-based cost aggregation and semi-global matching, followed by a left-right consistency check to eliminate errors in the
occluded regions.
While in \cite{LeCun2014} the CNN is used for comparing two image patches, our CNN is used to to compare rendered and observed images.

\noindent{\bf Analysis-by-synthesis}
has been a successful approach for many tasks in computer vision, such as object recognition \cite{Hejrati_CVPR_2014}, scene parsing \cite{Isola_SceneCollaging_2013}, viewpoint synthesis \cite{Hejrati_CVPR_2014}, material classification \cite{weinmann_synthesized_2014}, and gaze estimation \cite{sugano_synthesis_2014}. 
All these approaches use a forward model to synthesize some form of image, which is compared to observations.
Many works learn feature representation and compare in feature space.
For instance, in \cite{Hejrati_CVPR_2014} the analysis-by-synthesis strategy has been used for recognizing and reconstructing 3D objects in images.  
The forward model synthesizes visual templates defined on invariant features. 
Gall \etal \cite{GallRS_2008} propose an analysis-by-synthesis framework for motion capture and tracking.
It combines patch-based and region-based matching to track body parts.
Patch-based matching extracts correspondences between two successive frames for prediction and between the current image and a synthesized image for avoiding drift.
Recently, Brachmann \etal \cite{brachmann2014} achieved state-of-the-art results by adapting classical analysis-by-synthesis approach for 6D pose estimation of specific objects from a single RGB-D image. They use a new representation in form of a joint dense 3D object coordinate and object class labeling.
The major difference to our work, is that we learn to compare in the analysis-by-synthesis approach.
For the problem of 6D pose estimation, due to occlusion or complicated sensor noise, it can be difficult to compare the observation with the output of a rendered image of the object of interest in a particular pose. In this paper, we propose an approach, which draws on recent successes of CNNs. 
Different from aforementioned approaches, we model the posterior density of a particular object pose with a CNN that compares an observed and rendered image. The network is trained with the maximum likelihood paradigm.
One of the most closely related works is \cite{Kulkarni_2014}. They use a CNN as a part of probabilistic model.
The CNN is fed in a sequential manner, first with the rendered image, then with the observed image.
This produces two feature vectors, which are compared in the subsequence step, to give the probability of the observed image.
In contrast to \cite{Kulkarni_2014}, we jointly input the rendered and observed images into a CNN to produce an energy value.
The major difference is that our CNN is trained, while they take a pre-trained CNN as feature extractor.
\subsection{Review of the Pose Estimation Method \cite{brachmann2014} \label{sec:brachmann}}
We will now describe the system from \cite{brachmann2014} in detail, because it is of particular relevance for our method. Brachmann \etal \cite{brachmann2014} achieved state-of-the-art results by using a random forest \cite{breiman_2001} to obtain pixelwise dense predictions, which facilitate pose estimation. Each tree in their forest is trained to jointly predict to which object a pixel belongs to and, where it is located on the surface of this object. A tree outputs a soft segmentation image for each object with values between 0 and 1, indicating whether a pixel belongs to the object or not. The predictions of different trees are then combined to a single object probability. Additionally each tree outputs 3D object coordinates for each object and each pixel. The term object coordinates refers to the coordinates in the local coordinate system of the object. When estimating the pose of a particular object, Brachmann \etal \cite{brachmann2014} utilize the forest predictions in two ways:

Firstly, it is used to define an {\em energy function}, which is minimized to obtain the final pose. All aspects of the energy follow the analysis-by-synthesis principle. It is based on a pixelwise comparison between the predictions, the recorded depth values and rendered images of the object in the particular pose. In detail, three comparisons are done: (a) the rendered depth image of the object is compared to the recorded depth image; (b) the rendered image of object coordinates is compared to the predicted object coordinates; (c) the rendered segmentation mask of the object is compared to the predicted object class probability for the object. The pixelwise error inside the segmentation mask is aggregated and divided by the area of the mask. Robust error measures are used to deal with outliers.

Secondly, they use the forest predictions for an efficient {\em optimization scheme} to minimize the energy described above. It consists of two steps. The pixelwise object class probabilities are used inside the RANSAC pose estimation. In detail, sets of three pixels are sampled depending on the object class probability. For each set a pose hypothesis is calculated using the 3D-3D-correspondences between the camera coordinates, provided by the depth camera, and the object coordinates predicted by the forest. The best hypotheses, according to the energy function, are refined in a final step. Refinement is done by repeatedly determining inlier pixels in the rendered mask of the object, and again using the correspondences they provide to calculate a better pose. Finally. the pose with the lowest energy is taken as the final estimate.

In our work we build upon the framework of \cite{brachmann2014}. As in \cite{brachmann2014} we use the regression-classification random forest to obtain the predictions described above. We also use their optimization scheme, but replace the energy function with a novel one, based on a CNN, that is trained. The key difference is that while energy function in \cite{brachmann2014} has only a few parameters which can be trained via discriminative cross-validation procedure, the CNN has around 600K which we train with a maximum likelihood procedure. We show that this richness of parameters makes remarkable difference, and practical challenges such as occlusion and noise are much better dealt with. This approach will be described in the next section.

\section{Method}
\label{sec:methods}

\begin{figure}[!ht]
\begin{center}
\includegraphics [width=0.96\textwidth]{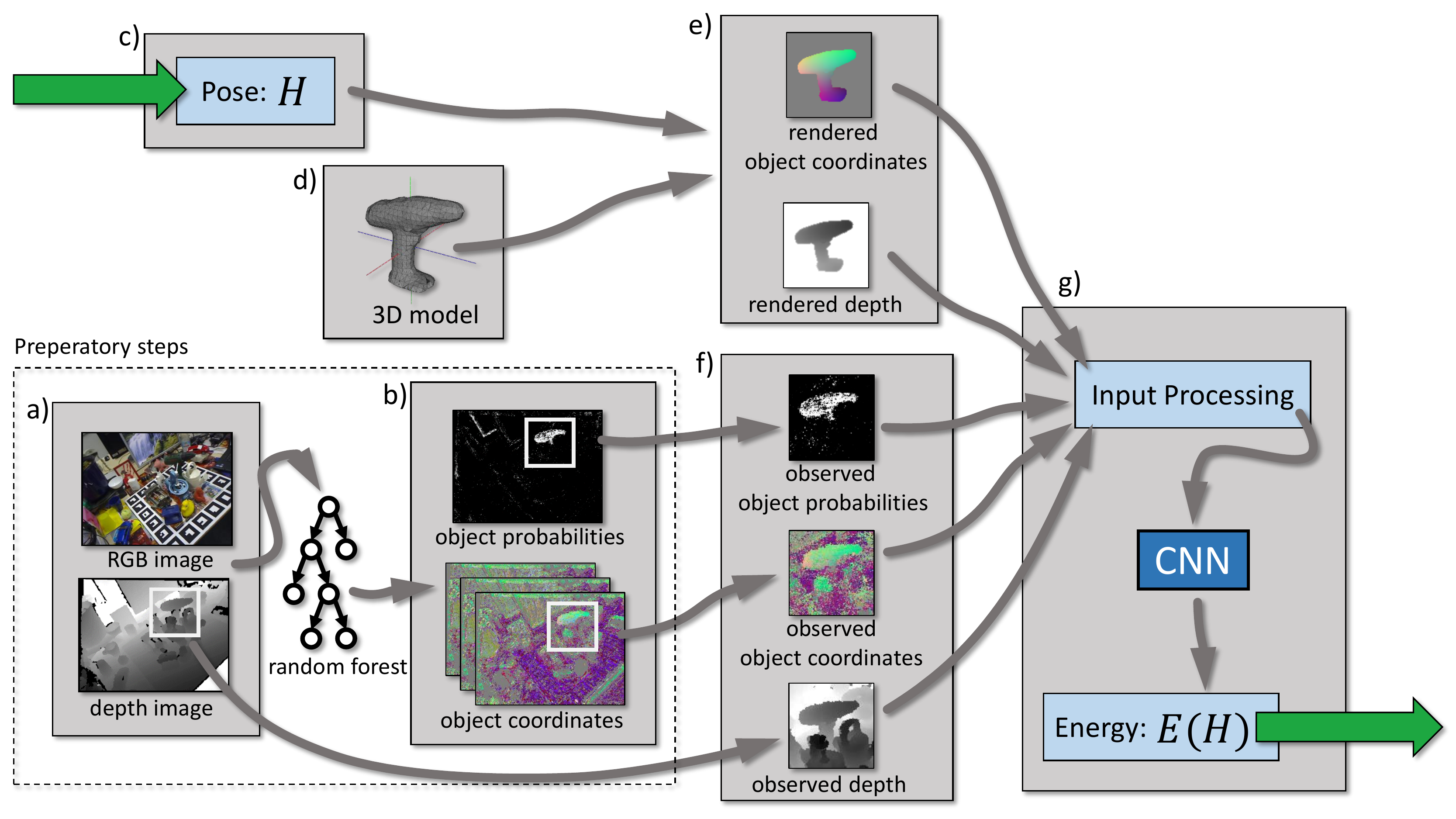}
\end{center}
 \caption{Our pipeline for the calculation of the energy function: Input and output are indicated by green arrows. The contents of the dashed box consists of preparatory steps, that have to be computed only once per image. (a) The RGB-D we will base our estimate on. The image is processed by a random forest to calculate predictions. (b) The predicted object probabilities and object coordinates. In the probability image bright pixels indicate a high probability. In the object coordinate images the 3D object coordinates are mapped to the RGB cube for visualization. There are multiple object coordinate images. Each one represents the prediction of one tree. The object coordinates are combined to a single image \cite{brachmann2014}. (c) The pose we want to calculate the energy for. (d) A 3D model of the object. (e) Images produced by rendering the 3D model in the input pose. We render an object coordinate image and a depth image. We only use cutouts around the object. (f) Images of equal size are cutout from the predicted object probabilities, object coordinates and from the recorded depth image. (g) Finally the rendered  and observed images are processed and fed into the CNN (Sec.~\ref{sec:convnet}). The single output of the CNN is our energy function.}
\label{fig:architecture}
\end{figure}

We will first give a description of the pose estimation task and introduce our terminology. Then we will describe our probabilistic model. The heart of this model is a CNN, which will be discussed subsequently. This is followed by a description of our maximum likelihood training procedure of the probabilistic model. Finally our inference procedure at test time is described. Fig.~\ref{fig:architecture} gives an overview of our testing pipeline. 

\subsection{The Pose Estimation Task \label{sec:task}}

We will now formally define the task of 6D pose estimation. Our goal is to estimate the pose $H$ of a rigid object\footnote{It should be noted that we assume the object to be present in the field of view, \ie we do not perform object recognition.} from a set of observations denoted by $\bx$, which will be discussed later. A pose describes the transformation from the local coordinate system of the object to the coordinate system of the camera. The local coordinate system has its origin in the center of the object. Each pose $H=(R,T)$ is a combination of two components. The rotational component $R$ is a $3 \times 3$ matrix describing the rotation around the center of the object. The translational component $T$ is a 3D vector corresponding to the position of the object center in the camera coordinate system. 

Let us now describe the observation $\bx$ that is used to estimate the object pose. We use RGB-D images as input. However, since we use the same random forest predictions as in \cite{brachmann2014}, the term observation or observed images will refer to two parts: (a)
the forest predictions as described in \cite{brachmann2014}, as well as (b) the recorded depth image. The reason for this simplified view is that the focus of our work lies on the modeling of the posterior density and aspects of the random forest prediction.

\subsection{Probabilistic Model}
We model the posterior distribution of the pose $H$ given the observations $\bx$ as a Gibbs distribution
\begin{equation} \label{eq:Gibbs}
	p(H| \bx ;\btheta)=\frac{\exp \big( -E(H,\bx;\btheta) \big)} 
	{\int \exp \big(-E (\hat{H},\bx;\btheta )  \big) d \hat{H}} \ ,
\end{equation}
where $ E\left(H,\bx;\btheta \right)$ is the so called energy function.  The energy function is a mapping from a pose $H$ and the observed images $\bx$ to a real number, parametrized by the vector $\btheta$. Note that using a Gibbs distribution to model the posterior is a common practice for conditional random fields (CRFs) \cite{Lafferty01}. However, the underlying energies are quite different. While in a CRF the energy function is a sum of potential functions, we implement it by using a CNN which directly outputs the energy value. The parameter vector $\btheta$ holds the weights of our CNN.

\subsection{Convolutional Neural Network \label{sec:convnet}}
In order to implement the mapping from a pose $H$ and the observed images $\bf x$ to an energy value we first render the object in pose $H$ to obtain rendered images ${\bf r}(H)$. Our CNN then compares $\bf x$ with ${\bf r}(H)$ and outputs a value $f\big( \bx, {\bf r}(H) ; \btheta \big)$. We define the energy function as
\begin{equation} \label{eq:ourEnergy}
E(H,\bx;\btheta)=f\big( \bx, {\bf r}(H) ; \btheta \big).
\end{equation}
Our network is trained to assign a low energy values when there is a large agreement between observed images and renderings and a high energy value when there is little agreement.
To perform the comparison we use a simple architecture, in which we feed all rendered and observed images as separate input channels into the CNN. 

Note that we consider only a square window around the center of the object with pose $H$. The width of the window is adjusted according to the size and distance of the object, as suggested by \cite{brachmann2014}. For performance reasons windows which are bigger than 100x100 pixels are down sampled to this size. We use in total six input channels for our network. Note that Fig.~\ref{fig:architecture} shows the images from which these six input channels are derived. 

\noindent One {\bf observed depth} channel and one {\bf rendered depth} channel that contain values in millimeters. They are normalized by subtracting the $z$ component of the object position according~to~$H$.

\noindent One {\bf rendered mask} channel of the object. Pixel values are either $+1$ for all pixels belonging to the object or $-1$ otherwise.

\noindent One {\bf depth mask} channel indicating whether a depth value was measured in the pixel. Again, pixel values are either $+1$ for all pixels where a depth was measured or $-1$ otherwise.

\noindent One {\bf probability} channel holding the combined pixel wise object probabilities from all trees. The values are re-scaled to lie between $-1$ and $+1$. 

\noindent One {\bf object coordinate} channel holding the pixel wise Euclidean distances between the rendered object coordinates and the predicted object coordinate from the tree giving the highest object probability for the respective pixel. We divide all values by the object diameter for normalization.

The $\tanh$ activation function is used after every convolution layer and after every fully connected layer. The first convolution layer $C_1$ consists 128 convolution kernels of size $3 \times 3 \times 6$. The second convolution layer $C_2$ consists of 128 kernels of size $3 \times 3 \times 128$, which is followed by a $2 \times 2$ max-pooling layer with stride 2 in each direction. The third convolution layer $C_3 $ is identical to $C_2$. The fourth convolution layer consists of $256$ kernels of size $3 \times 3 \times 128$. It is followed by a max-pooling operation over the remaining image size. The $256$ channels are further processed by two fully connected layers with $256$ neurons each and finally forwarded to a single output unit.

\subsection{Maximum Likelihood Training}
In training we want to find an optimal set of parameters $\btheta^*$ based on labeled training data $L=(\bx_1, H_1)\dots (\bx_n, H_n)$, where $\bx_i$ shall denote observations of the i-th training image and $H_i$ the corresponding ground truth pose. We apply the maximum likelihood paradigm and define  
\begin{equation} \label{eq:TrainingGoal}
	\btheta^*=\argmax_{\btheta} \sum_{i=1}^n \ln  p(H_i| \bx_i ;\btheta).
\end{equation}
In order to solve this optimization task we use stochastic gradient descent \cite{bottou1991stochastic}, which requires calculating the partial derivatives of the log likelihood for each training sample
\begin{equation} \label{eq:partial}
\begin{split}
	\frac
		{\partial }
		{\partial \theta_j} \ln  p (H_i| \bx_i ;\btheta)
	= 
	-\frac
		{\partial }
		{{\partial \theta_j}} E \left(H_i,\bx;\btheta \right) 
	\\ +\mathbb{E}\left[
	\frac
		{\partial }
		{{\partial \theta_j}} E\left(H,\bx_i;\btheta \right)
	\big| \bx_i ;\btheta \right]
\end{split}
\end{equation}
with respect to each parameter $\theta_j$. Here $\mathbb{E}[\cdot | \bx_i ;\btheta ]$ stands for the conditional expected value according to the posterior distribution $ p(H_i| \bx_i ;\btheta)$, parametrized by $\btheta$. While the partial derivatives of the energy function can be calculated by applying back propagation in our CNN, the expected value cannot be found in closed form. Therefore, we use the Metropolis algorithm \cite{metropolis} to approximate it, as discussed next.

\noindent{\bf Sampling.}
It is possible to approximate the expected value in Eq.~(\ref{eq:partial}) by a set of pose samples 
\begin{equation} \label{eq:sampling}
	\mathbb{E}\left[
	\frac
		{\partial }
		{{\partial \theta_j}} E\left(H,\bx_i;\btheta \right)
	\big| \bx_i ;\btheta \right]
	\approx
	\frac{1}{N}\sum_{k=1}^N
	\frac
		{\partial}
		{{\partial \theta_j}} E\left(H_k,\hat\bx;\btheta \right),
\end{equation}
where $H_1 \dots H_{N}$ are pose-samples drawn independently from the posterior $p(H| \bx ;\btheta)$ with the current parameters $\btheta$. We use the Metropolis algorithm \cite{metropolis} to generate these samples. It allows sampling from any distribution with a known density function that can be evaluated up to a constant factor. The algorithm generates a sequence of samples $H_t$ by repeating two steps:
\begin{enumerate}
  \item Draw a new proposed sample $H'$ according to a proposal distribution $Q(H'|H_t)$.
  \item Accept or reject the proposed sample according to an acceptance probability $A(H'|H_t)$. If the proposed sample is accepted set $H_{t+1}=H'$. If it is rejected set  $H_{t+1}=H_t$.
\end{enumerate}
The proposal distribution $Q(H'|H_t)$ has to be symmetric, i.e. $Q(H'|H_t)=Q(H_t|H')$. Our particular proposal distribution will be described in detail in the next section. The acceptance probability is in our case defined as
\begin{equation} \label{eq:acceptance}
	A(H'|H_t)=\min \left(1,
	\frac{p(H'| \bx ;\btheta)}
		{p(H_t| \bx ;\btheta)}
	\right),
\end{equation}
meaning that  whenever the posterior density $p(H'| \bx ;\btheta)$ at the proposed sample is greater than the posterior density  $(H_t| \bx ;\btheta)$ at the current sample, the proposed sample will automatically be accepted. If this is not the case it will be accepted with the probability $p(H'| \bx ;\btheta)/p(H_t| \bx ;\btheta)$.

\noindent{\bf Proposal Distribution.}
A common choice for the proposal distribution is a normal distribution centered at the current sample. In our case this is not possible because the rotational component of the pose lives on the manifold $SO(3)$, i.e. the group of rotations. We define $Q(H'|H_t)$ implicitly by describing a sampling procedure and ensuring that it is symmetric. The translational component $T'$ of the proposed sample is directly drawn from a 3D isotropic normal distribution $\mathcal{N}(T_t,{ \Sigma}_T)$ centered at the translational component $T_t$ of the current sample $H_t$. The rotational component $R'$ of the proposed sample $H'$ is generated by applying a random rotation $\hat{R}$ to the rotational component $R_t$ of the current sample:
$R' = \hat{R} R_t$.

We calculate $\hat{R}$ as the rotation matrix corresponding to an Euler vector \footnote{A 3D vector  represents a rotation. The direction of the vector describes the axis of the rotation and the length corresponds to the angle.} ${\bf e}$, which is drawn from a 3D zero centered isotropic normal distribution ${\bf e} \sim \mathcal{N}({\bf 0},{ \Sigma}_R)$.

\noindent{\bf Initialization and Burn-in-phase.}
When the Metropolis algorithm is initialized in an area with low density it requires more iterations to provide a fair approximation of the expected value. To find a good initialization we run our inference procedure (described in the next section) using the current parameter set. We then perform the Metropolis algorithm for a total of 130 iterations, disregarding the samples from the first 30 iterations which are considered as burn-in-phase.  


\subsection{Inference Procedure \label{sec:inference}}
During test time we aim at finding the MAP estimate, i.e. the pose maximizing our posterior density as given in Eq.~(\ref{eq:Gibbs}). Since the denominator in Eq.~(\ref{eq:Gibbs}) is constant for any given observation $\bx$, finding the MAP estimate is equivalent to minimizing our energy function. To achieve this, we utilize the optimization scheme from \cite{brachmann2014}, but replace their energy function with ours. 

\section{Experiments}
\label{sec:exp}

In the following we compare our approach to the state-of-the-art method of Brachmann \etal in \cite{brachmann2014} for two different datasets. We first describe some implementation details of the competitor and introduce the datasets. After that we describe details of our training procedure, and finally present quantitative and qualitative  comparison. We will see that we achieve considerable improvements for both datasets. Additionally, we observe that our CNN generalizes from a single training object to a set of 11 test objects, with large variability in appearance and geometry.

\subsection{Datasets, Competitors, Evaluation Protocol}

\noindent {\bf Datasets.} We use two datasets featuring heavy occlusion. The first dataset was created by Brachmann \etal. \cite{brachmann2014} by annotating the ground truth poses for eight partially occluded objects in images taken from the dataset of Hinterstoisser \etal \cite{hinterstoisser2012accv}. We will refer to this dataset as the \emph{occlusion dataset} from \cite{brachmann2014} and \cite{hinterstoisser2012accv}. It includes a total of $8992$ test cases (images with different annotation), which are used for testing. We choose this dataset because it is more challenging than the original dataset from \cite{hinterstoisser2012accv}, on which  \cite{brachmann2014} already achieves an average of 98.3\% correctly estimated poses.

The second dataset was introduced by Krull \etal in \cite{krull2014}.
It provides six annotated RGB-D sequences of three different objects and consists of a total of $3187$ images. We use three of the sequences for training and the other three (a total of $1715$ test images) for testing.

\noindent {\bf Evaluation Protocol.} We use the evaluation procedure as described in \cite{brachmann2014}. This means we calculate the percentage of correctly predicted poses for each sequence. As in \cite{hinterstoisser2012accv} we calculate the average distance between the 3D model vertices under the estimated pose and under the ground truth pose. A pose is considered correct, when the average distance is below 10\% of the object diameter.

\noindent {\bf Competitors.} We compare our method to the one presented in \cite{brachmann2014}. For doing so we needed to re-implement this method\footnote{Our re-implementation is identical up to small details, which we discussed with the authors of \cite{brachmann2014}.}.  We observed that our re-implementation gives on average slightly superior results. In the following, we mostly report two numbers, those of our re-implementation and those of the method of \cite{brachmann2014}, reported in \cite{brachmann2014} or \cite{krull2014}. For completeness we additionally provide the numbers from LineMOD \cite{hinterstoisser2012accv} as reported in \cite{brachmann2014}.

\begin{figure}[!ht]
\begin{center}
\includegraphics [width=0.8\textwidth]{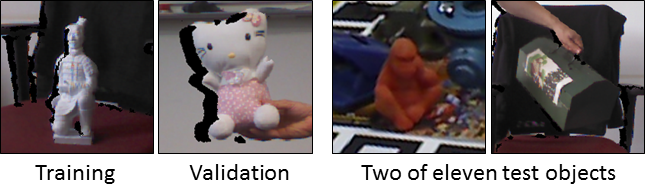}
\end{center}
   \caption{Images from one of our training-testing configurations: the \emph{Samurai\_1} sequence is used for training,  the \emph{Cat\_1} for validation. Sequences of all objects are used for testing. Note, the Objects are of vastly different shape and appearance.}
\label{fig:trainValidateTest}
\end{figure}


\subsection{Training Procedure}
\noindent {\bf Random Forests.}
We used different random forests for training and testing on both datasets. The forests were kindly provided to us by the authors of \cite{brachmann2014}. 

\noindent {\bf CNN.}
We trained three CNNs, each time using only a single object from the dataset provided by Krull \etal in \cite{krull2014}. The sequences \emph{Toolbox\_1}, \emph{Cat\_1}, and \emph{Samurai\_1} served as training sets - see Fig.~\ref{fig:trainValidateTest}. The first $100$ frames from \emph{Samurai\_1} were removed in order to obtain a high percentage of frames with occlusion. Our validation set consists of $100$ randomly selected frames from the \emph{Cat\_1} sequence, or the \emph{Samurai\_1} sequence (in the case where \emph{Cat\_1} was used as training set).
The weights of the CNN were randomly initialized. Before training, the random weights of the last layer were multiplied by factor 1000, in order to cover a greater range of possible energy values.
After every 5th iteration of stochastic gradient descent, we perform inference on the validation set and  adjust the learning rate. The learning rate at step $t$ was proportional to $\gamma_t= \gamma_0 / (1+\gamma_0 \lambda t)$ \cite{bottou-tricks-2012}, with $\gamma_0=10$ and $\lambda=0.5$.
After training we pick the set of weights which achieved the highest percentage of correctly estimates poses on the validation set. We use the criterion from \cite{hinterstoisser2012accv} to classify a pose as correct. One training cycle consisting of five steps of stochastic gradient descent and validation took\footnote{We used an \emph{Intel(R) Core (TM) i7-3820 CPU at 3.60GHz} with GeForce GTX 660 GPU. The \emph{Cat\_1} sequence was used for training and 100 random frames from \emph{Samurai\_1} for validation.} 9min 46sec (2min 27sec + 7min 19sec).
Further details on our training procedure can be found in the supplementary material. 


\begin{figure}[!ht]
\begin{center}
\includegraphics [width=0.8\textwidth]{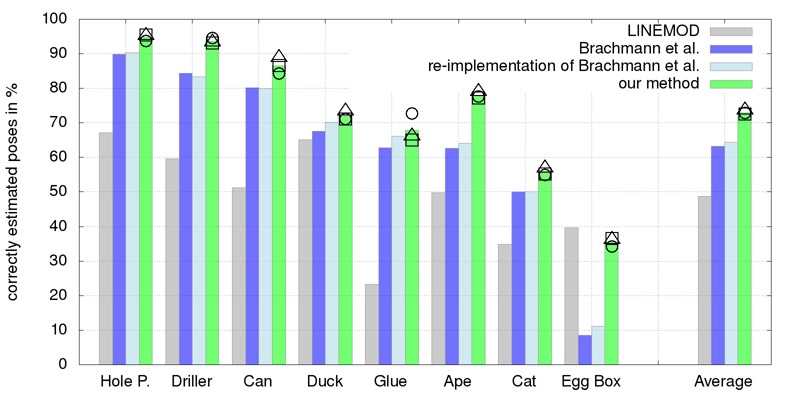}
\end{center}
   \caption{Quantitative comparison of our method against the results of \cite{brachmann2014} and LineMOD \cite{hinterstoisser2012accv} on the \emph{Occlusion Dataset} from  \cite{brachmann2014} and \cite{hinterstoisser2012accv}. \emph{Circles}, \emph{Squares}, and \emph{Triangles} indicate the individual performance of CNNs trained with \emph{Tool Box}, \emph{Cat}, and \emph{Samurai} respectively. The green bars indicate the average result. Averaged over all test and training objects we obtain the correct pose in {\bf 72.98\%} of cases, in contrast to 63.24\% for \cite{brachmann2014} and 48.84\% for LineMOD \cite{hinterstoisser2012accv}. A table with the the detailed numbers can be found in the supplementary material.}
\label{fig:comp-eccv14}
\end{figure}

\subsection{Comparison}

\begin{figure}[t]
\begin{center}
\includegraphics [width=0.8\textwidth]{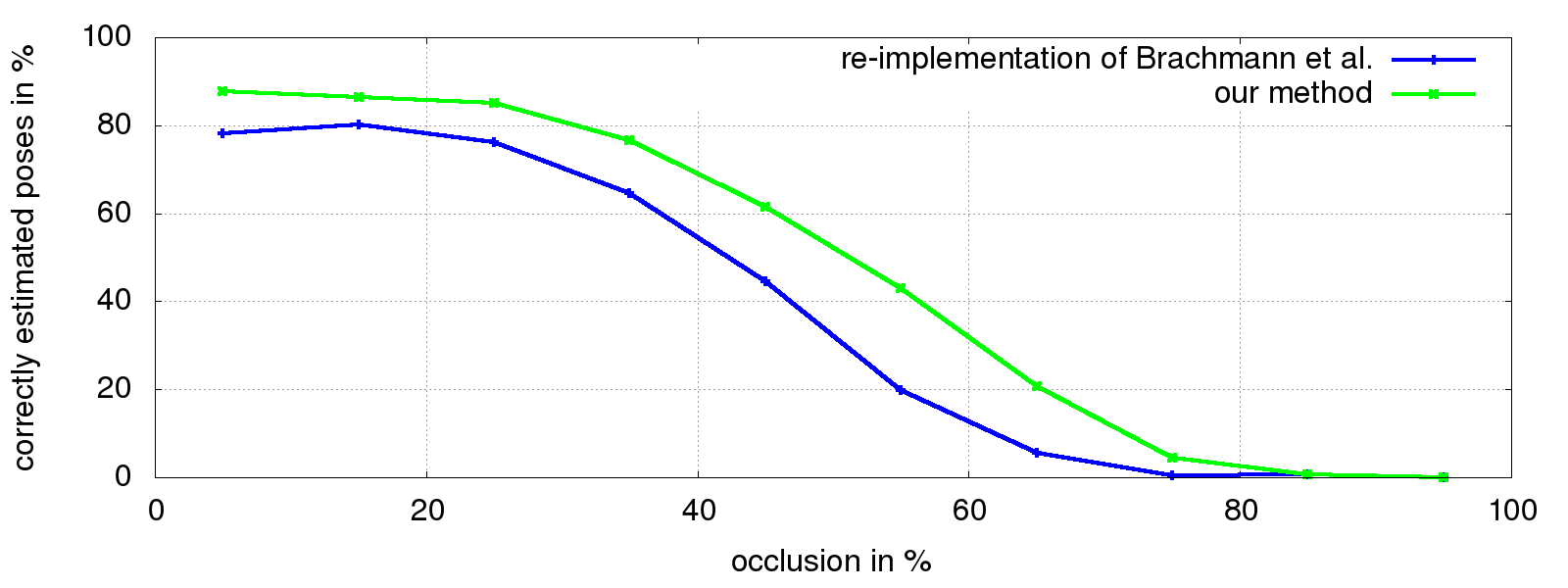}
\end{center}
   \caption{The percentage of correctly estimated poses for all test cases of the \emph{occlusion dataset} from \cite{brachmann2014} and \cite{hinterstoisser2012accv}, as a function of the level of occlusion. For this we divided the test cases into bins according to the amount of occlusion, using a bin width of 10\%. (See details of this procedure in the supplementary material.) We compare our method (using the CNN trained with the \emph{Samurai} object) to our re-implementation of \cite{brachmann2014}. We achieve improvements of over 20\% for occlusion levels between 50\% and 60\%.}
\label{fig:comp-occlusion-eccv14}
\end{figure}

\begin{figure}[!ht]
\begin{center}
\includegraphics [width=0.7\textwidth]{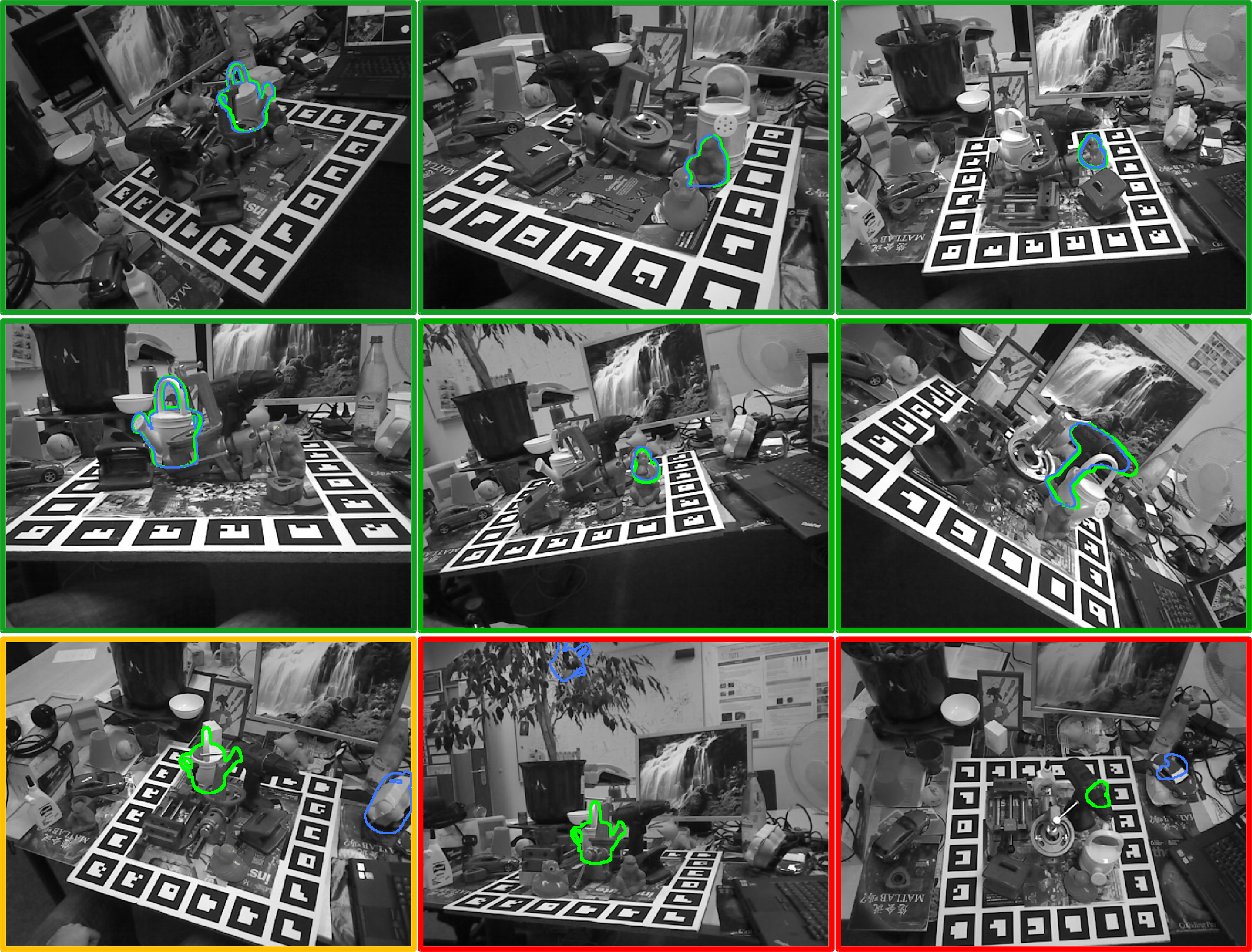}
\end{center}
   \caption{Qualitative results of our method on the \emph{occlusion dataset} from \cite{brachmann2014} and \cite{hinterstoisser2012accv}. Here green and blue silhouettes correspond to the ground truth and our estimate, respectively. The test images depicted with a green frame show correct estimates. Images with orange and red frame show incorrect estimates. The image with an orange frame shows a case where the energy of the ground truth pose, according to Eq.~(\ref{eq:ourEnergy}), is lower than the energy of the estimated pose. In this case a better pose may be found with an improved optimization scheme.}
\label{fig:pose-results}
\end{figure}

\noindent{\bf \emph{Occlusion Dataset} from  \cite{brachmann2014} and \cite{hinterstoisser2012accv}.}
Quantitative results for this dataset are shown in Fig.~\ref{fig:comp-eccv14}, for all individual test and training objects. Considering the average over all objects we achieve an improvement of up to ${9.23\%}$ compared to our re-implementation of \cite{brachmann2014} and {\bf 10.4\%} compared to the reported values in \cite{brachmann2014}. Some qualitative results are illustrated in Fig.~\ref{fig:pose-results}. In Fig.~\ref{fig:comp-occlusion-eccv14} we show another comparison of our method with respect to \cite{brachmann2014}. It illustrates that we achieve the biggest gain for occlusion percentage between $50\%$ and $60\%$. 

\begin{figure}[!ht]
\begin{center}
\includegraphics [width=0.8\textwidth]{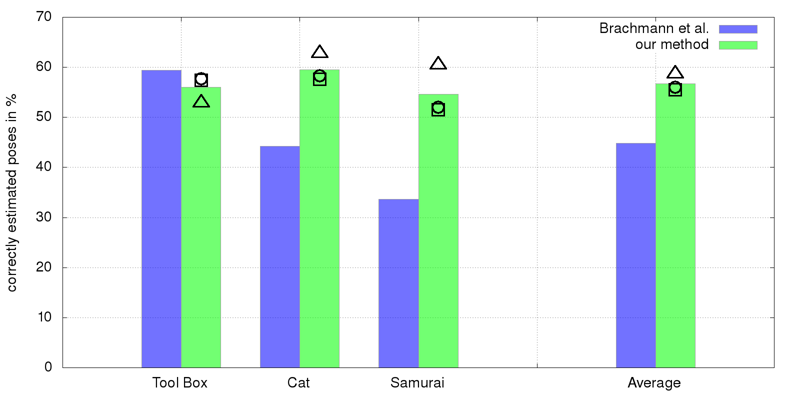}
\end{center}
   \caption{Comparison of our method on the dataset of Krull \etal., against the results of \cite{brachmann2014}.  \emph{Circles}, \emph{Squares}, and \emph{Triangles} indicate the individual performance of CNNs trained with \emph{Tool Box}, \emph{Cat}, and \emph{Samurai} respectively. The green bars indicate the average result. We report $56.02\%$, $59.56\%$, and $54.65\%$ correctly estimated poses for \emph{Tool Box}, \emph{Cat}, and \emph{Samurai} respectively. Averaged over all test and training objects we achieve {\bf 56.74\%}.}
\label{fig:comp-accv14}
\end{figure}

\begin{figure}[!ht]
\begin{center}
\includegraphics [width=0.7\textwidth]{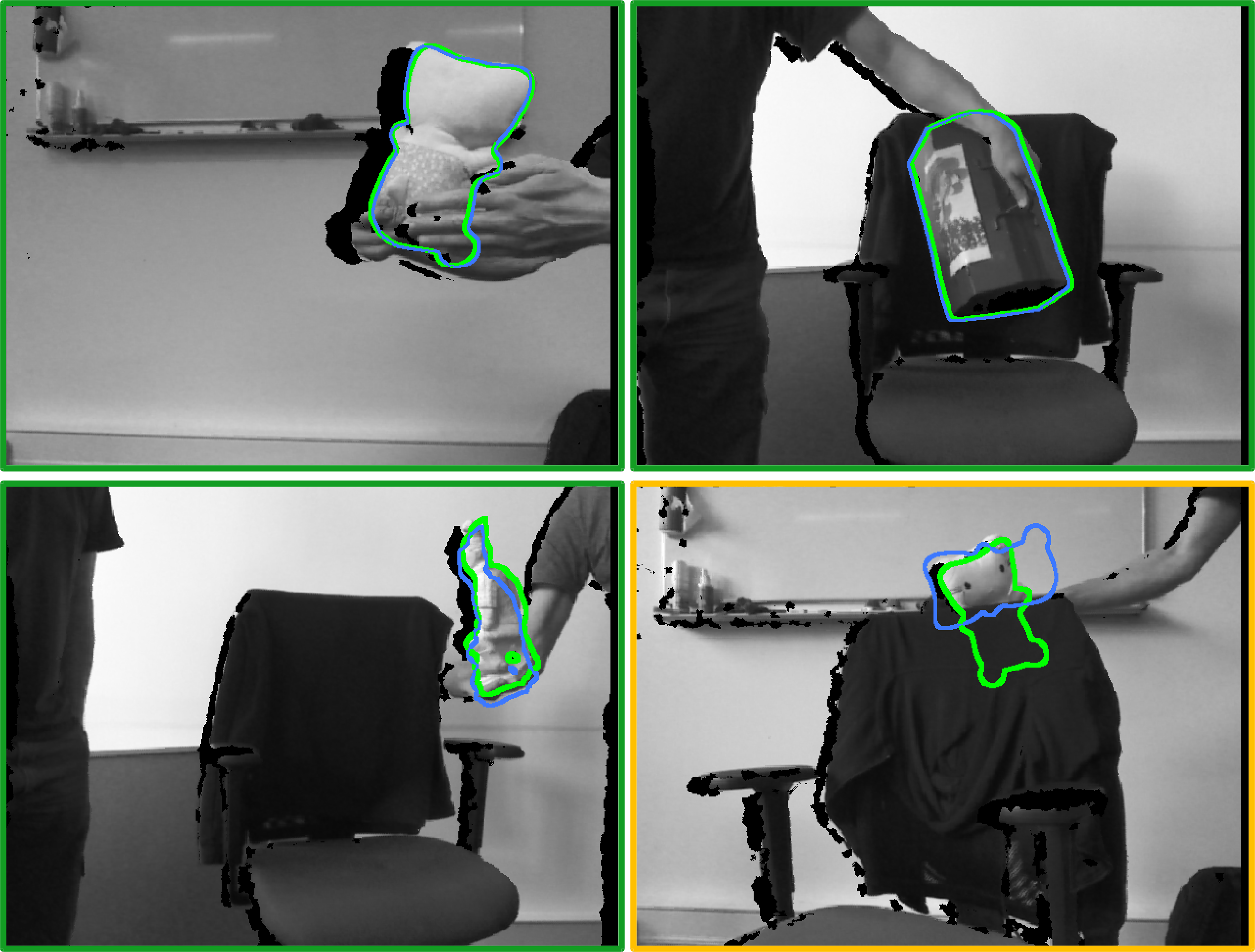}
\end{center}
   \caption{
Qualitative results of our method on the test cases from the dataset introduced in \cite{krull2014}: Green frames correspond to correctly estimated poses according to the criteria from \cite{hinterstoisser2012accv}. Orange frames correspond to incorrectly estimated poses with a lower energy at the ground truth than at the estimated pose.}
\label{fig:pose-accv-results}
\end{figure}

\noindent{\bf Dataset of Krull \etal}
For this dataset we observe similar results as with the previous dataset. Since the other sequences were used in training and validation, we evaluated only with the \emph{Toolbox\_2}, \emph{Cat\_2}, and \emph{Samurai\_2} sequences. When averaged over all objects we achieve an improvement of {\bf 10.97\%} compared to the results of \cite{brachmann2014}. The quantitative results can be found in Fig.~\ref{fig:comp-accv14}, and a few qualitative results are shown in Fig.~\ref{fig:pose-accv-results}. 

\noindent{\bf Discussion of Failure Cases.}
The failure cases which are framed red in Fig.~\ref{fig:pose-results} have to be considered as failure of our learned energy function. However, the failure cases framed orange still exhibit a lower energy at the ground truth pose than at the estimate.
This indicates a failure of the optimization scheme. It should be investigated in which case the correct pose can be found using an alternative optimization scheme. 
In the dataset introduced by Krull \etal our accuracy for the \emph{Tool Box} sequences is below the one of our competitor (see Fig.~\ref{fig:comp-accv14}).
We attribute this to the fact that the \emph{Tool Box} is the biggest object and most strongly affected by the down sampling schema described in Sec.~\ref{sec:convnet}.

\section{Conclusion}
\label{sec:con}

We have presented a model for the posterior distribution in 6D pose estimation, which uses a CNN to map rendered and observed images to an energy value. We train the CNN based on the maximum likelihood paradigm. It has been demonstrated, that training on a single object is sufficient and the CNN is able to generalize to different objects and backgrounds. Our system has been evaluated on two datasets featuring heavy occlusion. By using our energy as objective function for pose estimation, we were able to achieve considerable improvements compared to the best previously published results.

Our approach is not restricted to the feature channels  and even the application we demonstrated. The architecture of the CNN can in principle be applied to any kind of observed and rendered image. We think it would be worth investigating if the approach could be applied to other scenarios. An example could be pose estimation from pure RGB without recorded depth image and a forest to calculate features. Pose estimation for object classes could also benefit from our approach. Considering the recent success of CNNs in recognition  \cite{agrawal_nn_2014,oquab_cnn_2014} it might be possible for a CNN to learn to compare observed images to renderings of an idealized model representing an object class instead of an instance. Our approach is not limited to comparing images of the same kind, as for example rendered and observed depth images. Instead, it could learn to asses the plausibility of the shading in an observed RGB by comparing it to a rendered depth image, which can be more easily produced than a realistic RGB rendering.

An interesting future line of research could be to train a CNN to predict pose updates from observed and rendered images. This could replace the refinement step and might improve the results.
\newpage

{\small
\bibliographystyle{ieee}
\bibliography{myLib}
}

\end{document}